\documentclass[10pt,twocolumn,letterpaper]{article}
\usepackage{cvpr}
\usepackage[table]{xcolor}
\usepackage{booktabs}
\usepackage{nicematrix}
\usepackage{array}
\newcolumntype{P}{>{\columncolor{black!5}}c}
\newcolumntype{M}{>{\columncolor{blue!6}}c}
\newcolumntype{R}{>{\columncolor{red!7}}c}
\usepackage{arydshln}
\usepackage[T1]{fontenc}
\usepackage[utf8]{inputenc}
\usepackage{graphicx}
\usepackage{amsmath,amssymb}
\usepackage{booktabs,tabularx}
\usepackage{xcolor}
\usepackage{tikz}
\usetikzlibrary{arrows.meta,positioning,calc}
\usepackage{algorithm}
\usepackage{algpseudocode}

\definecolor{cvprblue}{rgb}{0.21,0.49,0.74}
\definecolor{policyblue}{RGB}{227,239,250}
\definecolor{monitorgreen}{RGB}{230,242,230}
\definecolor{agentorange}{RGB}{253,237,217}
\usepackage[pagebackref,breaklinks,colorlinks,allcolors=cvprblue]{hyperref}
\usepackage[capitalize,noabbrev]{cleveref}

\def\confName{CVPR}
\def\confYear{2027}

\title{Reasoning on Call: Efficient Agentic Robot Control\\
for Long-Horizon Robot Manipulation}

\author{
    Zilang Cen\textsuperscript{1,2,3},
    Chao Tang\textsuperscript{3},
    Fangcheng Liu\textsuperscript{4},
    Weishi Mi\textsuperscript{3},
    Wei Hao\textsuperscript{3},
    JiaYu Xu\textsuperscript{3}
    \\[0.3em]
    Dehao Huang\textsuperscript{2,3,5},
    Yehui Tang\textsuperscript{4},
    Xiaohui Cui\textsuperscript{1},
    Tingguang Li\textsuperscript{3},
    and Xiaoyuan Zhang\textsuperscript{2,*}
    \\[0.7em]
    {\small
        \textsuperscript{1}Wuhan University
        \quad
        \textsuperscript{2}Zhongguancun Academy
        \quad
        \textsuperscript{3}Samsung Robotics eXperience
    }
    \\[0.25em]
    {\small
        \textsuperscript{4}Samsung R\&D Institute China--Beijing
        \quad
        \textsuperscript{5}Southern University of Science and Technology
    }
}

\begin{document}
\maketitle

\begin{abstract}
Large multimodal models can support long-horizon robot manipulation, but frequent model calls incur substantial latency and token costs. The central challenge is determining when agent assistance or intervention is needed and whether errors require correcting robot execution, subtask planning, or both. To address this problem, we propose \textbf{RoC} (Reasoning on Call), an efficient framework that coordinates a low-level policy and a high-level reasoning agent through an online \textbf{task-state monitor}. The monitor combines temporal visual observations with ordered interaction memory to jointly estimate progress within the active subtask and predict its successor. These predictions guide event-triggered agent review for coordinated execution repair and plan revision. Verified outcomes govern memory updates and subtask transitions, allowing routine execution to remain with the policy while reasoning focuses on decisions requiring additional assessment. On RMBench, RoC improves task success rate by 24.0 percentage points and reduces token consumption by 65.3\% compared with the strongest agent-based baseline. It also outperforms the strongest policy-based baseline by 15.1 percentage points in task success rate. Real-world experiments further validate its effectiveness across three long-horizon manipulation tasks. Overall, RoC supports efficient agent intervention and coordinated recovery across execution and planning for long-horizon robot manipulation.
\end{abstract}

\section{Introduction}\label{sec:intro}

Long-horizon robot manipulation requires both reliable execution and effective planning. A robot must carry out individual subtasks, assess their outcomes, and determine which goal to pursue next. Visuomotor policies, such as VLAs~\cite{pmlr-v305-black25a} and WAMs~\cite{ye2026world}, provide action execution capability, but successful long-horizon manipulation also requires coordinating these actions with task-level decisions.

Recently, agentic robot control approaches such as GPT-as-Policy~\cite{su2026astra} and Harness VLA~\cite{zhang2026harnessvla} introduce multimodal reasoning agents to interpret task context, coordinate subtasks, and guide recovery. However, repeated agent calls incur substantial token costs and inference latency~\cite{sun2026call,li2026spotter}. These costs accumulate as the agent repeatedly processes visual observations and execution history. When execution must wait for its decisions, agent intervention becomes a bottleneck even during routine behavior that the policy can handle independently. The challenge is therefore to determine \textit{when} agent intervention is needed and \textit{how} it should guide execution repair and plan revision, without repeatedly querying the agent itself.

\begin{figure}[t]
  \centering
  \includegraphics[width=\linewidth]{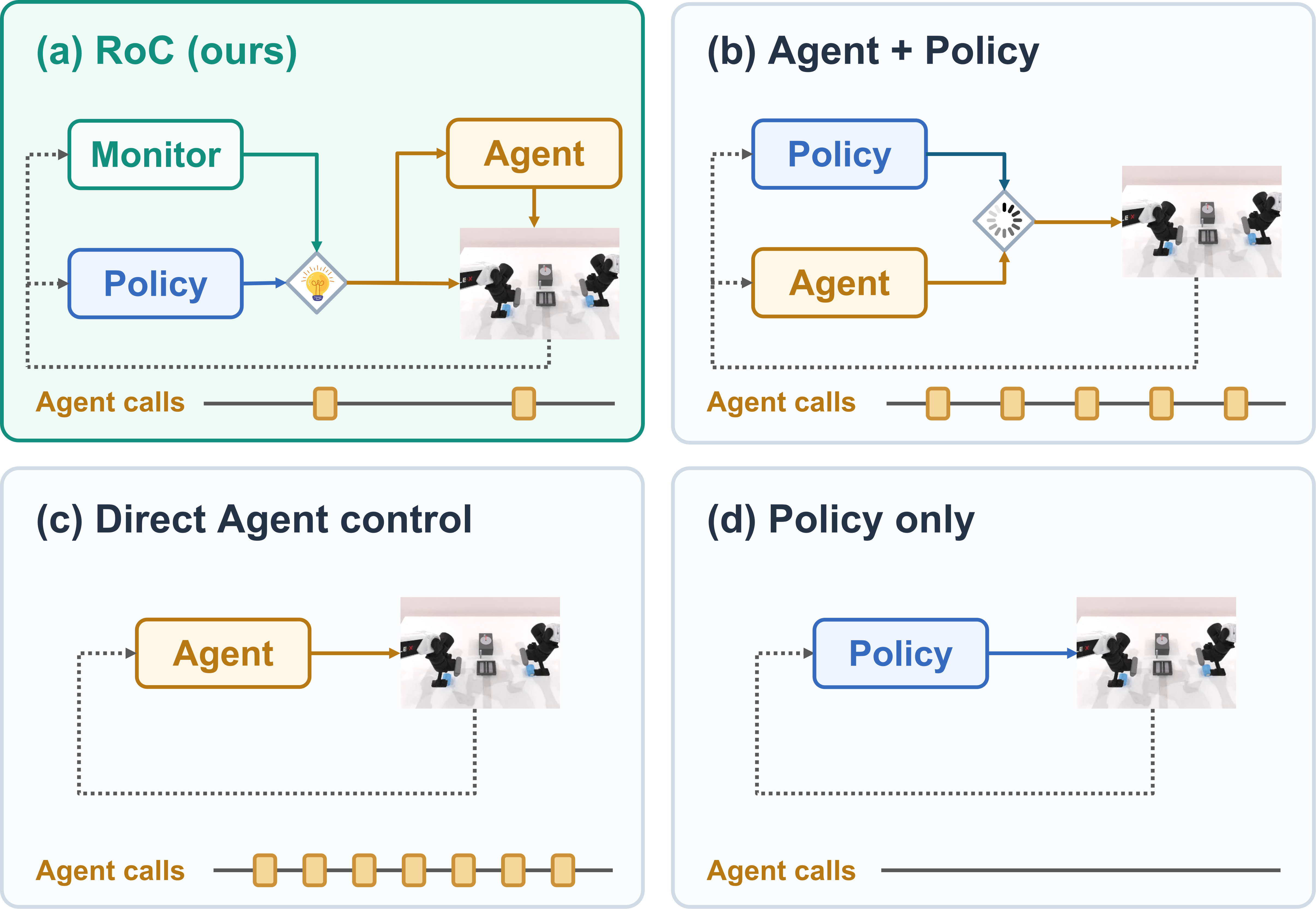}
  \caption{\textbf{Efficient agentic robot control with RoC.} (a) Monitor-triggered agent intervention reduces its token overhead while supporting execution repair and plan revision. (b,c) Frequent agent calls incur repeated reasoning costs, whereas (d) policy-only control lacks agent assistance. Gold markers denote agent calls.}
    \vspace{-0.5cm}
  \label{fig:teaser}
\end{figure}

To address this challenge, we propose \textbf{RoC} (Reasoning on Call), an efficient framework that coordinates a low-level policy and a reasoning agent through an online \textbf{task-state monitor}. The monitor integrates stage-aware visual context with ordered interaction memory to jointly estimate active subtask progress, predict transition requests, and propose next goals. Progress anomalies, transition requests, and critical contact events in candidate actions selectively trigger agent review. Using shared task and execution context, the agent can authorize continued policy execution, shorten or refine candidate actions, take over subtask execution, or revise the proposed next goal. Verified outcomes govern memory updates and subtask transitions. This design allows routine execution to remain with the policy while directing agent reasoning toward decisions that require further assessment or correction.

We evaluate RoC on both simulated and real-world tasks, assessing task success and token consumption. On RMBench, RoC improves task success rate by 24.0 percentage points and reduces token consumption by 65.3\% compared with the strongest agent-based baseline. It also outperforms the strongest policy-based baseline by 15.1 percentage points in task success rate. Three long-horizon tasks on the SO-ARM101 platform further demonstrate the effectiveness of RoC in real-world settings.

To summarize, our main contributions are twofold:
\begin{itemize}
    \item We propose RoC, an efficient agentic robot control framework for long-horizon manipulation that couples monitor-triggered agent review with coordinated execution repair and plan revision.
    \item We develop an online task-state monitor that serves as a plug-in module to guide selective agent intervention, reducing redundant agent calls and reasoning overhead.
\end{itemize}

\section{Related Work}
\label{sec:related}

\subsection{Long-Horizon Manipulation}
Long-horizon manipulation requires robots to execute interdependent subtasks while retaining historical observations and interaction outcomes. Policy-based approaches address this by incorporating task history into action generation and subtask selection. MEM~\cite{torne2026memmultiscaleembodiedmemory} and MemoryWAM~\cite{yang2026memorywamefficientworldaction} equip policies with short- and long-term memory, while MemER~\cite{sridhar2025memerscalingmemoryrobot} and PAL~\cite{liu2026planfirstactlater} use learned hierarchical policies to retrieve historical observations or organize them into executable subtasks. These methods improve memory-conditioned decisions, but recovery still depends on the learned policies' ability to correct failed interactions and inappropriate subtask choices. RoC complements these policies with selective agent intervention for execution repair and plan revision, expanding recovery capabilities while preserving efficient routine execution.

\subsection{Agentic Robot Control}
Recent advances in general-purpose multimodal models have enabled agentic robot control through visual understanding and semantic reasoning. Specifically, RoboICL~\cite{liu2026roboiclembodiedincontextlearning} generates actions using demonstrations and interaction memory, while CaP-X~\cite{fu2026capx} synthesizes executable robot-control programs. GPT-as-Policy~\cite{su2026astra} explores both direct agent control and a hybrid mode in which the agent reviews and corrects actions proposed by $\pi_{0.5}$. Harness VLA~\cite{zhang2026harnessvla} combines agent reasoning with learned manipulation skills by composing VLA and analytic primitives for task orchestration and recovery.

However, frequent agent calls introduce latency and token costs. Cocurrent work reduces this overhead through selective assistance~\cite{sun2026call,li2026spotter}, feedback-driven replanning~\cite{lei2026longhorizonembodiedagentstoolaligned,chen2026astrarobomme}, and learned monitors for progress, failures, and task transitions~\cite{huang2026stageguard,ajith2026robomonitor,bhardwaj2026progressprobes}. Detecting a problem or predicting a task transition does not, by itself, determine whether recovery requires correcting execution, revising the plan, or both. RoC makes this coordination explicit: the agent jointly reviews the current execution and proposed next goal, allowing execution repair and plan revision within the same intervention.




\subsection{Reward and Progress Models in Robotics}
Reward and progress models provide intermediate feedback for evaluating robot execution and improving policies~\cite{zhang2025rewind}. For multi-stage tasks, VICtoR~\cite{hung2025victor} identifies task stages and evaluates progress within each stage. SARM~\cite{chen2026sarm} jointly predicts task stage and fine-grained progress, using estimated rewards to filter and reweight demonstrations for behavior cloning. Robometer~\cite{liang2026robometer} combines within-trajectory progress supervision with cross-trajectory preference learning, leveraging successful and failed executions for reward modeling, policy optimization, and failure detection. Building on Robometer, RoC's online task-state monitor jointly estimates subtask progress, predicts transition requests, and proposes next goals, guiding selective agent intervention for coordinated execution repair and plan revision.

\section{Method}
\label{sec:method}

RoC follows a monitoring--execution--review loop. At initialization, the online task-state monitor proposes the first subtask from the task instruction and initial observation, using empty subtask and interaction-memory inputs. The reasoning agent confirms this proposal before policy execution begins. During execution, the monitor estimates the active subtask progress and predicts transition requests and next goals. Transition requests, progress anomalies, and critical contact events trigger agent review for coordinated execution repair and plan revision. Verified outcomes and approved transitions govern memory updates and task progression. We formulate the component interfaces in Sec.~\ref{sec:formulation}, then describe the online task-state monitor in Sec.~\ref{sec:monitor} and the agent intervention mechanism in Sec.~\ref{sec:events}.


\begin{figure*}[!ht]
  \centering
  \includegraphics[width=0.85\linewidth]{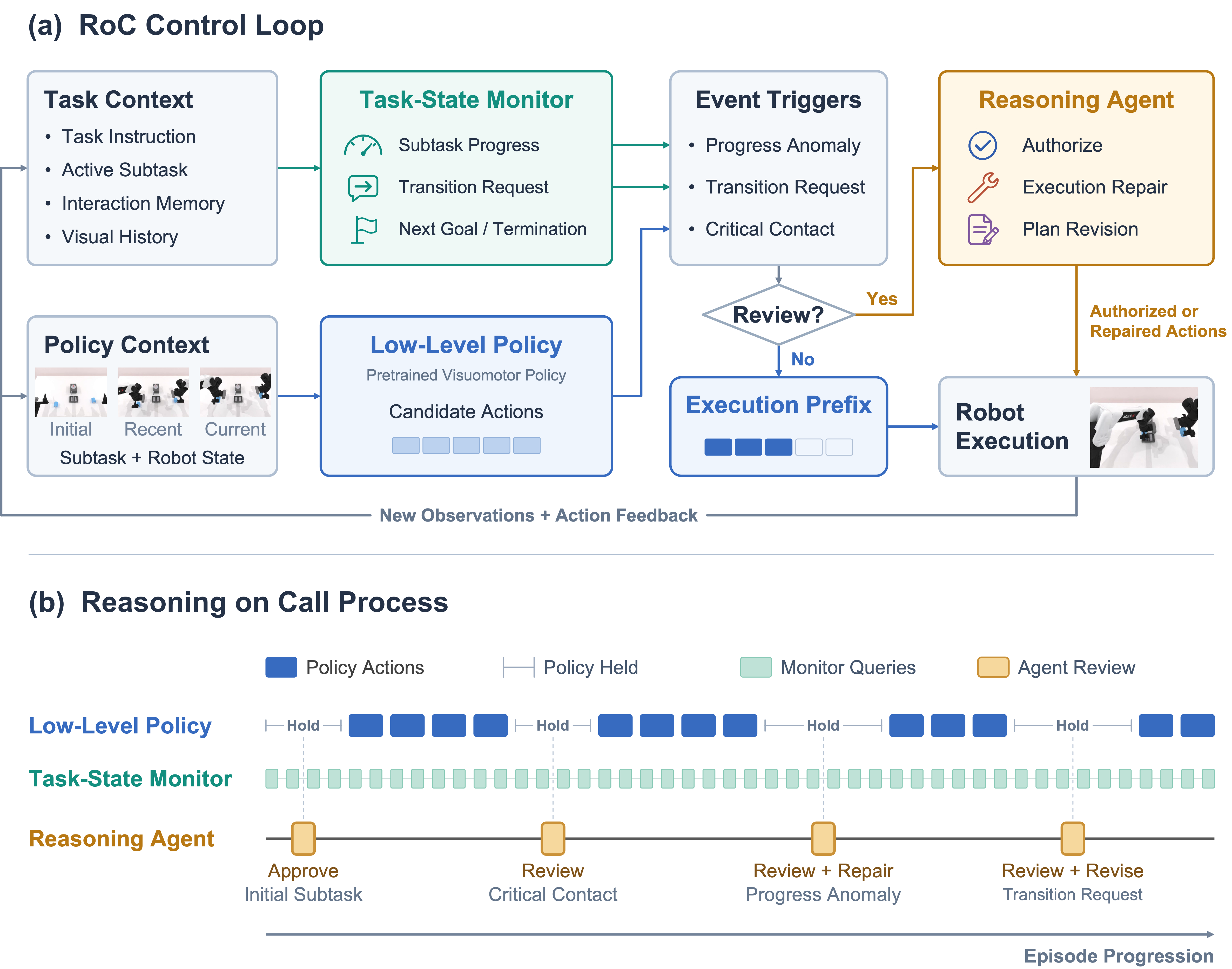}
  \caption{\textbf{RoC control loop and reasoning on call.} (a) The online task-state monitor estimates subtask progress, predicts transition requests, and proposes the next subtask or task termination. The low-level policy proposes an action chunk. RoC executes its prefix unless a progress anomaly, transition request, or critical contact event triggers agent review. During review, the agent can authorize continued execution or intervene through execution repair, plan revision, or both. (b) The agent approves the initial subtask and is called selectively thereafter. Monitor queries continue during policy holds, so routine execution can proceed between agent calls.}
    \vspace{-0.3cm}
  \label{fig:pipeline}
\end{figure*}

\subsection{Problem Formulation}
\label{sec:formulation}

Given a language instruction $g$ specifying a task, the robot receives an RGB observation $o_t$ and proprioceptive state $q_t$ at global control step $t$. RoC maintains visual history $\mathcal{H}_t$ (including $o_t$), ordered interaction memory $\mathcal{M}_t$ containing verified records, and an active subtask $\ell_k$. Here, $k$ indexes subtask instances independently of $t$.

The online task-state monitor estimates progress within the active subtask and proposes subtask transitions:
\begin{equation}
    (p_t,s_t,\widetilde{\ell}_t)
    =F_\phi(g,\ell_k,\mathcal{M}_t,\mathcal{H}_t)
\end{equation}
where $p_t\in[0,1]$ is the progress estimate of $\ell_k$, $s_t$ is the transition-request probability, and $\widetilde{\ell}_t$ is a candidate successor or task-termination proposal. Progress estimates are used to detect execution anomalies, while $s_t$ determines whether to request a transition. The candidate $\widetilde{\ell}_t$ specifies the proposed next goal.



The low-level policy proposes an action chunk at each control step:
\begin{equation}
    A_t \sim \pi_\theta(\cdot \mid \ell_k, q_t, \mathcal{H}_t^\pi)
\end{equation}
where $A_t=(a_{t,1},\ldots,a_{t,H})$ and $\mathcal{H}_t^\pi$ denotes the observations selected from $\mathcal{H}_t$ for policy inputs. Before executing an initial segment of $A_t$, termed the execution prefix, RoC evaluates three triggers for agent review: progress anomalies, transition requests, and critical contact events in $A_t$. The corresponding triggering mechanism is defined in Sec.~\ref{sec:events}. If no review is triggered, RoC executes the prefix and obtains new observations. Otherwise, it withholds the affected actions while the agent reviews the candidate actions and any requested subtask transition.

\subsection{Progress Monitoring and Subtask Prediction}
\label{sec:monitor}

\paragraph{Interaction Memory and Visual Context.}
The monitor conditions on $g$, $\ell_k$, $\mathcal{M}_t$, and observations selected from $\mathcal{H}_t$. Each entry in $\mathcal{M}_t$ contains a subtask instruction, executed actions, an outcome description, and timestamps corresponding to supporting observations in $\mathcal{H}_t$. Records follow execution order, with repeated instructions retained as distinct subtask instances. Their generation and commitment are described in Sec.~\ref{sec:events}. 

\begin{figure*}[h]
  \centering
  \includegraphics[width=\linewidth]{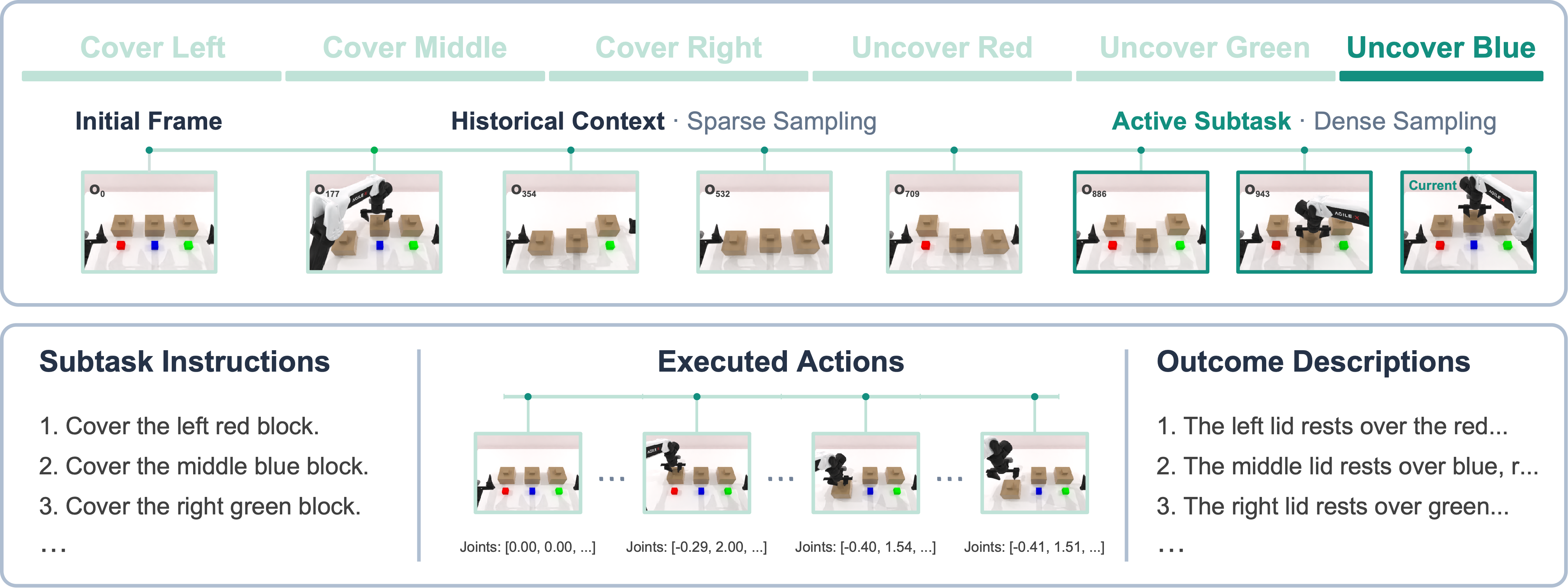}
  \caption{Illustration of visual history $\mathcal{H}_t$ (top) and ordered interaction memory $\mathcal{M}_t$ (bottom).}
    \vspace{-0.5cm}
  \label{fig:htmt}
\end{figure*}

We select visual context from $\mathcal{H}_t$ using stage-aware sampling. Sparse observations uniformly sampled from the initial time step to the start of the current subtask provide a persistent scene reference and historical context. Denser observations are uniformly sampled from the start of $\ell_k$ to time $t$, capturing local execution changes. This balances historical coverage and local detail under a fixed visual input budget, while $\mathcal{M}_t$ preserves interaction outcomes that sparse observations may miss, as illustrated in Figure~\ref{fig:htmt}.

\paragraph{Online Task-State Monitor.}
We initialize $F_\phi$ from Robometer~\cite{liang2026robometer}, a general-purpose robotic reward model. Its pretraining combines progress supervision with trajectory comparisons involving successful, suboptimal, and failed executions, providing a pretrained basis for assessing execution progress.

Visual observations sampled from $\mathcal{H}_t$, together with $g$, $\ell_k$, and the ordered subtask list recorded in $\mathcal{M}_t$, form a shared context for the monitor's predictions. Specifically, learnable progress tokens and densely sampled observations are causally interleaved. The contextual hidden state of the current observation's progress token feeds both the progress head and the transition-request head. The progress head predicts a categorical distribution $\mathbf{p}_t$ over ten discrete progress values in $[0,1]$, whose expectation gives the scalar estimate $p_t$. This estimate describes the relative execution stage within $\ell_k$ and is used for progress-anomaly detection.
 
The transition-request head predicts $s_t$ directly from the same contextual representation. Although the two heads share their underlying features, they have different supervision and control roles: $p_t$ identifies abnormal progress patterns, while $s_t$ requests a change of subtask. An autoregressive language generation branch proposes $\widetilde{\ell}_t$ or task termination from the shared context. The request head therefore predicts whether a transition should be reviewed, while the generation branch specifies which subtask should follow. During execution, the generation branch is activated only when $s_t$ exceeds the transition threshold; otherwise, successor generation is skipped. Progress and transition-request predictions are read from context positions preceding the successor text, preventing the target instruction from leaking into these predictions during training. Completion is verified by the reasoning agent during review, as described in Sec.~\ref{sec:events}.

\paragraph{Training Objective.}
Subtask transitions are temporally sparse, so uniform sampling yields far more training examples within subtasks than near their boundaries. This imbalance can make accurate transition timing difficult to learn. We therefore halve the interval between sampled training timesteps around annotated subtask boundaries, doubling the local sampling density and increasing the monitor's exposure to states immediately before and after a transition. We jointly supervise progress estimation, transition requests, and next goal generation:
\begin{equation}
\label{eq:monitorloss}
\begin{split}
    \mathcal{L}_{\mathrm{monitor}} ={}&
    \mathcal{L}_{\mathrm{AR}}
    + \operatorname{CE}(y_t,\mathbf{p}_t) + \operatorname{BCE}(s_t^\star,s_t)
\end{split}
\end{equation}
where $\mathcal{L}_{\mathrm{AR}}$ supervises autoregressive successor generation, $y_t$ is the target progress distribution within $\ell_k$, and $s_t^\star$ is the transition-request label. Training inputs contain only observations and verified interaction records available up to time $t$.

\subsection{Event-Triggered Agent Intervention}
\label{sec:events}


\paragraph{Event Triggers.}
We define three event types.
(1) Transition events occur when $s_t$ exceeds the request threshold. The candidate $\widetilde{\ell}_t$ specifies the next goal or task termination to be reviewed; it does not independently trigger a transition.
(2) Progress events occur when $p_t$ exhibits significant declines, repeated oscillations, or stagnation within the same subtask instance $\ell_k$.
(3) Critical contact events occur before grasp closure or placement release in $A_t$, prompting checks of object identity, relative geometry, and interaction prerequisites.

\paragraph{Agent Intervention and Control Handoff.}
\label{sec:repair}
Using shared task and execution context, the agent jointly determines whether execution repair, plan revision, or both are needed. \textbf{Execution repair} supports three modes: (1) prefix shortening, which executes fewer actions from $A_t$ before obtaining new observations; (2) local refinement, which adjusts end-effector poses or gripper commands; and (3) subtask takeover, in which the agent directly controls the robot until the subtask outcome is verified. \textbf{Plan revision} supports two modes: (1) transition rejection, which retains an unfinished $\ell_k$; and (2) successor revision, which replaces an invalid $\widetilde{\ell}_t$ with an appropriate next subtask. If neither correction is needed, the agent authorizes continued policy execution.

Following review or repair, the agent checks observations and executed actions to verify subtask completion, generating an outcome description when completion is confirmed. RoC stops executing the completed $\ell_k$ and retains its outcome while reviewing the successor. Once the successor or task termination is approved, RoC updates the ordered interaction memory $\mathcal{M}_t$ with the verified interaction record and commits the transition. If the successor is rejected, RoC replans without repeating the completed interaction. Policy commands are suspended during agent takeover. Before policy execution resumes, RoC discards stale actions and generates a new action chunk from fresh observations and the confirmed subtask.




\section{Experiments}
\label{sec:experiments}

\subsection{Experimental Setup}
\paragraph{Environments and Tasks.}
In simulation, we evaluate RoC on all nine tasks in RMBench~\cite{chen2026rmbench}, where an $M(1)$ task requires recalling a key past observation (e.g., the initial position of an object), while an $M(n)$ task requires tracking multiple past interactions (e.g., previous attempts or completed operations) to guide subsequent actions. For real-robot evaluation, we instantiate RoC on the 5-DoF SO-ARM101 platform, with a fixed external RGB camera and a wrist-mounted RGB camera. We evaluate three long-horizon manipulation tasks, Drawer Sorting, Cover and Reveal, and Tabletop Cleanup. 

\paragraph{Baselines.}
We organize the baselines into two groups and evaluate them separately. Agent-based baselines include direct agent control and hybrid control in GPT-as-Policy~\cite{su2026astra}. These baselines retain their original configurations with GPT-6 Astra (xhigh), while RoC uses GPT-6.1 Sol (xhigh) for agent intervention. Seven policy-based baselines include standard policies and memory-augmented policies, as listed in Table~\ref{tab:memory_comparison}.

\paragraph{Evaluation Protocol.}
We measure task success rate (SR) and average token consumption per episode (Tok.). In simulation, comparisons with agent-based methods use 30 independent trials per task, while comparisons with policy-based methods use 100 independent trials per task. We report per-task results and averages across the nine tasks. Each real-robot task is evaluated over 10 independent trials, with success defined as completing all required subtasks in the specified order and achieving the target final state.

\subsection{Implementation Details}
In simulation, we train the online task-state monitor for one epoch across all nine RMBench tasks. RoC supports any pretrained low-level policy in theory. In our experiments, we initialize the policy from pretrained FastWAM~\cite{yuan2026fastwam}. The policy receives the current subtask instruction, robot state and three visual observations: the initial frame $o_0$, a recent frame from the active subtask, and the current frame $o_t$. 

During inference, the monitor operates at 7.10\, Hz without subtask generation and 0.53\, Hz when generating a subtask, with an overall measured throughput of 2.77 Hz. The low-level policy uses 50 diffusion denoising steps to predict a 32-step action chunk. During agent review, the reasoning agent can approve an execution prefix of 1–32 steps or apply local corrections over 1–5 steps. For progress anomaly detection, we first smooth progress estimates using a median filter and then detect decline, oscillation, or stagnation within a 64-step lookback window. The decline threshold is set to 0.2. Oscillation is detected when the progress range exceeds 0.25 with more than three reversals in the direction of meaningful progress changes, while stagnation is identified when the progress range remains within 0.1.


\subsection{Simulation Evaluation}
\paragraph{Comparison with Agent-Based Methods.}
Table~\ref{tab:rmbench_comparison} reports task success and token consumption for agent-based methods. RoC achieves an average success rate of 99.6\%, outperforming GPT-as-Policy and Direct Control by 24.0 and 27.7 percentage points, while reducing token consumption by 65.3\% and 84.4\%, respectively. These gains stem from event-triggered intervention guided by the online task-state monitor, enabling flexible agent intervention at critical moments. GPT-as-Policy repeatedly invokes the agent to review policy proposals, introducing reasoning overhead even during routine execution. Frequent review, however, does not guarantee accurate assessment of execution outcomes or successful correction~\cite{su2026astra}. RoC guides agent review using learned subtask-progress estimates, proposed next goals, and verified interaction records. These provide explicit task context for deciding whether to continue execution, repair an action, or revise the plan. 

\begin{figure}[t]
  \centering
  \includegraphics[width=\linewidth]{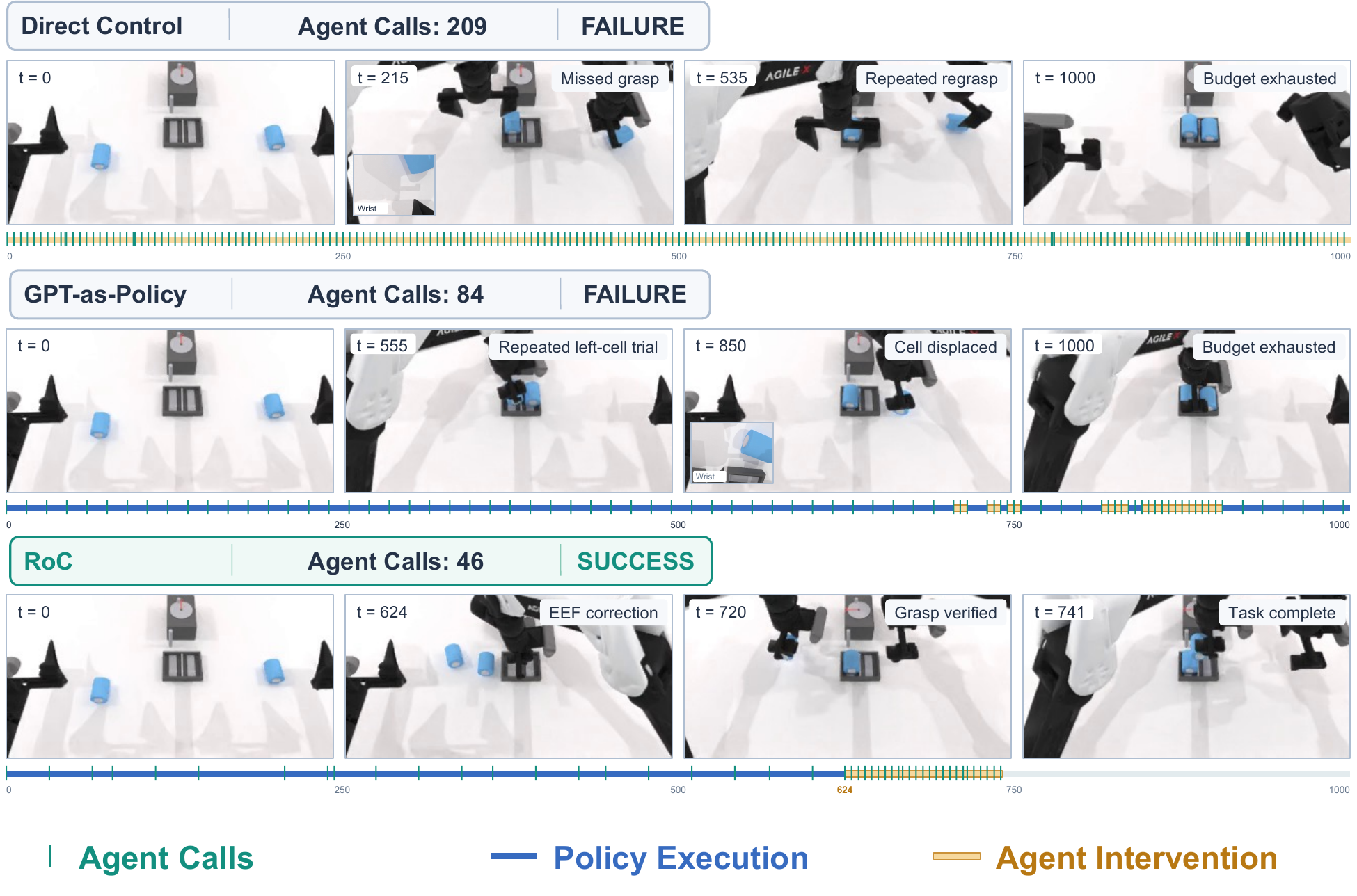}
  \caption{Qualitative comparison of agent-based methods. Vertical markers indicate agent calls, while blue and yellow segments denote policy execution and agent intervention, respectively.}
  \label{fig:agent_compare}
\end{figure}


As illustrated in Figure~\ref{fig:agent_compare}, Direct Control repeatedly attempts failed grasps, wwhile GPT-as-Policy exhausts its budget on repeated failed attempts to manipulate the battery. In contrast, RoC's design enables targeted agent intervention for execution repair, requiring only 46 agent calls compared with 84 and 209 for GPT-as-Policy and Direct Control, respectively. These results highlight the efficiency advantage of the proposed RoC framework.

\paragraph{Comparison with Policy-Based Methods.}

Table~\ref{tab:memory_comparison} compares RoC with standard and memory-augmented policies. RoC achieves an average success rate of 98.1\%, outperforming MemoryWAM, the strongest policy-based baseline, by 15.1 percentage points. While memory-based policies benefit from historical context, they still rely on learned action generation to handle execution failures and task progression. With selective agent intervention, RoC provides enhanced capabilities for correcting execution errors and revising subsequent goals with limited reasoning overhead.




\begin{table}[t]
  \centering
  \caption{Comparison with agent-based methods on RMBench. SR denotes success rate (\%), and Tok. denotes average token consumption per episode (in units of $10^4$ tokens).}
  \label{tab:rmbench_comparison}
  \scriptsize
  \resizebox{\columnwidth}{!}{%
    \begin{tabular}{@{}lcccccc@{}}
      \toprule
      & \multicolumn{2}{c}{{Direct Control}} & \multicolumn{2}{c}{{GPT-as-Policy}~\cite{su2026astra}} & \multicolumn{2}{c}{\textbf{RoC}} \\
      \cmidrule(lr){2-3} \cmidrule(lr){4-5} \cmidrule(l){6-7}
      {Task} & {SR} & {Tok.} & {SR} & {Tok.} & {SR} & {Tok.} \\
      \midrule
      \multicolumn{1}{@{}c}{\textit{$M(1)$ Tasks}} \\
      Observe and Pick Up & 86.7 & 403.9 & 73.3 & 409.0 & 96.7 & 391.0 \\
      Rearrange Blocks & 43.3 & 1678.7 & 66.7 & 572.6 & 100.0 & 83.3 \\
      Put Back Block & 96.7 & 485.6 & 90.0 & 308.8 & 100.0 & 73.7 \\
      Swap Blocks & 66.7 & 1908.6 & 70.0 & 1035.2 & 100.0 & 265.4 \\
      Swap T & 100.0 & 849.6 & 100.0 & 297.8 & 100.0 & 171.2 \\
      \midrule
      \multicolumn{1}{@{}c}{\textit{$M(n)$ Tasks}} \\
      Battery Try & 73.3 & 1894.1 & 53.3 & 832.5 & 100.0 & 126.2 \\
      Blocks Ranking Try & 100.0 & 3115.0 & 96.7 & 1387.2 & 100.0 & 944.3 \\
      Cover Blocks & 13.3 & 4513.0 & 30.0 & 2123.9 & 100.0 & 131.1 \\
      Press Button & 66.7 & 1757.6 & 100.0 & 505.5 & 100.0 & 408.7 \\
      \midrule
      \textbf{Overall Avg.} & 71.9 & 1845.1 & 75.6 & 830.3 & \textbf{99.6} & \textbf{288.3} \\
      \bottomrule
    \end{tabular}%
  }
\end{table}


\begin{table*}[t]
\centering
\caption{Comparison with policy-based methods on RMBench. Success rates are reported as percentages.}
\label{tab:memory_comparison}
\scriptsize
\resizebox{\textwidth}{!}{%
\begin{tabular}{@{}lcccccccc@{}}
\toprule
& \multicolumn{3}{c}{Policy Only}
& \multicolumn{4}{c}{Memory-Based}
& \\
\cmidrule(lr){2-4}\cmidrule(lr){5-8}
\textbf{Task}
& X-VLA~\cite{zheng2026x}
& FastWAM~\cite{yuan2026fastwam}
& $\pi_{0.5}$~\cite{pmlr-v305-black25a}
& MemoryVLA~\cite{shi2026memoryvla}
& Mem-0~\cite{chen2026rmbench}
& EventVLA~\cite{yang2026eventvla}
& MemoryWAM~\cite{yang2026memorywamefficientworldaction}
& \textbf{RoC} \\
\midrule
\multicolumn{1}{@{}c}{\textit{$M(1)$ Tasks}} \\
Observe and Pick Up & 9.0 & 0.0 & 9.0 & 2.0 & 4.0 & 21.0 & 27.0 & 95.0 \\
Rearrange Blocks & 13.0 & 0.0 & 13.0 & 53.0 & 89.0 & 96.0 & 100.0 & 100.0 \\
Put Back Block & 18.0 & 0.0 & 11.0 & 81.0 & 90.0 & 95.0 & 100.0 & 100.0 \\
Swap Blocks & 16.0 & 0.0 & 24.0 & 76.0 & 67.0 & 96.0 & 100.0 & 100.0 \\
Swap T & 3.0 & 7.0 & 15.0 & 9.0 & 14.0 & 87.0 & 94.0 & 99.0 \\
\midrule
\multicolumn{1}{@{}c}{\textit{$M(n)$ Tasks}} \\
Battery Try & 26.0 & 20.0 & 16.0 & 33.0 & 28.0 & 35.0 & 41.0 & 99.0 \\
Blocks Ranking Try & 1.0 & 26.0 & 6.0 & 53.0 & 18.0 & 81.0 & 100.0 & 98.0 \\
Cover Blocks & 2.0 & 0.0 & 0.0 & 69.0 & 68.0 & 97.0 & 98.0 & 99.0 \\
Press Button & 0.0 & 0.0 & 0.0 & 0.0 & 0.0 & 3.0 & 87.0 & 93.0 \\
\midrule
\textbf{Overall Avg.} & 9.8 & 5.9 & 10.4 & 41.7 & 42.0 & 67.8 & 83.0 & \textbf{98.1} \\
\bottomrule
\end{tabular}%
}
\end{table*}

\subsection{Ablation Studies}
\label{sec:ablations}


\begin{table}[t]
\centering
\caption{Ablation of online task-state monitoring and event-triggered agent intervention on RMBench. Success rates are reported as percentages.}
\label{tab:roc_ablation}
\scriptsize
 \resizebox{\columnwidth}{!}{%
\begin{tabular}{@{}lcccc@{}}
\toprule
\textbf{Task} & \shortstack{Policy\\ Only}  & \shortstack{Policy\\+ History} & \shortstack{Policy + History\\+ Monitor} & \shortstack{RoC \\ (Full)} \\
\midrule
\multicolumn{1}{@{}c}{\textit{$M(1)$ Tasks}} \\ 
Observe and Pick Up & 0.0 & 4.0 & 8.0 & 95.0 \\
Rearrange Blocks & 0.0 & 17.0 & 94.0 & 100.0 \\
Put Back Block & 0.0 & 13.0 & 93.0 & 100.0 \\
Swap Blocks & 0.0 & 7.0 & 94.0 & 100.0 \\
Swap T & 7.0 & 12.0 & 80.0 & 99.0 \\
\midrule
\multicolumn{1}{@{}c}{\textit{$M(n)$ Tasks}} \\
Battery Try & 20.0 & 35.0 & 85.0 & 99.0 \\
Blocks Ranking Try & 26.0 & 38.0 & 35.0 & 98.0 \\
Cover Blocks & 0.0 & 4.0 & 39.0 & 99.0 \\
Press Button & 0.0 & 2.0 & 45.0 & 93.0 \\
\midrule
\textbf{Overall Avg.} & \textbf{5.9} & \textbf{14.7} & \textbf{63.7} & \textbf{98.1} \\
\bottomrule
\end{tabular}
}
\end{table}

Table~\ref{tab:roc_ablation} evaluates the incremental contributions of visual history, online task-state monitoring, and event-triggered agent intervention. Policy Only uses the original FastWAM. Both Policy Only and Policy + History follow the full task instruction, with the latter additionally using visual history. Adding the monitor enables execution guided by predicted subtask instructions, while the full RoC further incorporates event-triggered agent intervention. Visual history increases average success from 5.9\% to 14.7\%, showing its benefits  but remain insufficient for reliable long-horizon execution. More importantly, the monitor raises success to 63.7\%, with substantial gains on Rearrange Blocks and Put Back Block.


However, task-state monitoring alone cannot reliably resolve failed execution or incorrect subtask transitions. This limitation is particularly evident in the $M(n)$ tasks, where the monitor-equipped policy achieves an average success rate of 51.0\%. With event-triggered agent intervention, RoC increases the overall success rate by 34.4 percentage points to 98.1\%, while improving the $M(n)$ average to 97.3\%. These gains highlight the complementary roles of monitoring and agent intervention: the monitor provides task-state estimates and transition proposals, while the reasoning agent verifies execution outcomes and coordinates execution repair with plan revision when necessary.

\begin{figure}[h]
  \centering
  \includegraphics[width=\linewidth]{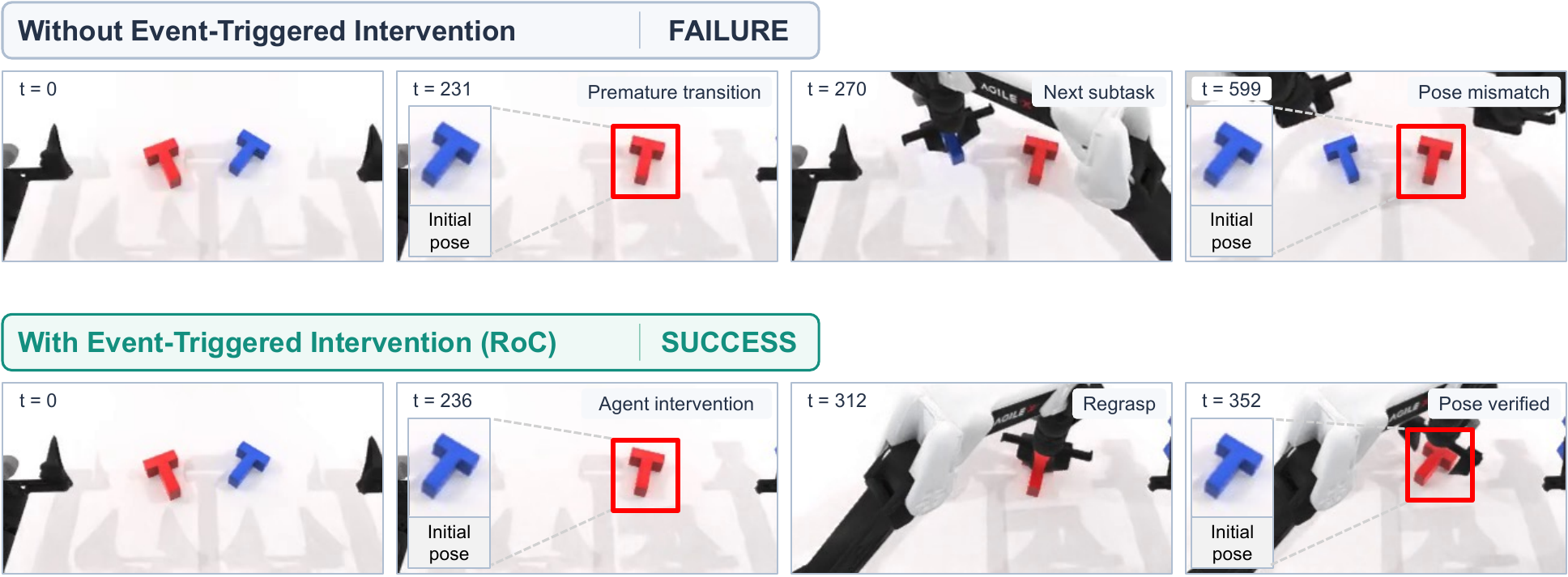}
  \caption{Qualitative comparison on RMBench without event-triggered agent intervention (top) and with the full RoC framework (bottom).}
  \label{fig:policy_compare}
\end{figure}

To further illustrate the role of event-triggered agent intervention in execution repair, Figure~\ref{fig:policy_compare} presents a representative failure case. Without agent intervention, the system prematurely advances to the next subtask despite an incorrect object orientation, ultimately resulting in a pose mismatch. In contrast, RoC triggers agent intervention to correct the grasp and verify the target pose before proceeding. This example further highlights the importance of timely error correction in preventing execution failures from propagating across subtasks.

\subsection{Real-Robot Evaluation}

\begin{figure*}[h]
  \centering
  \includegraphics[width=\linewidth]{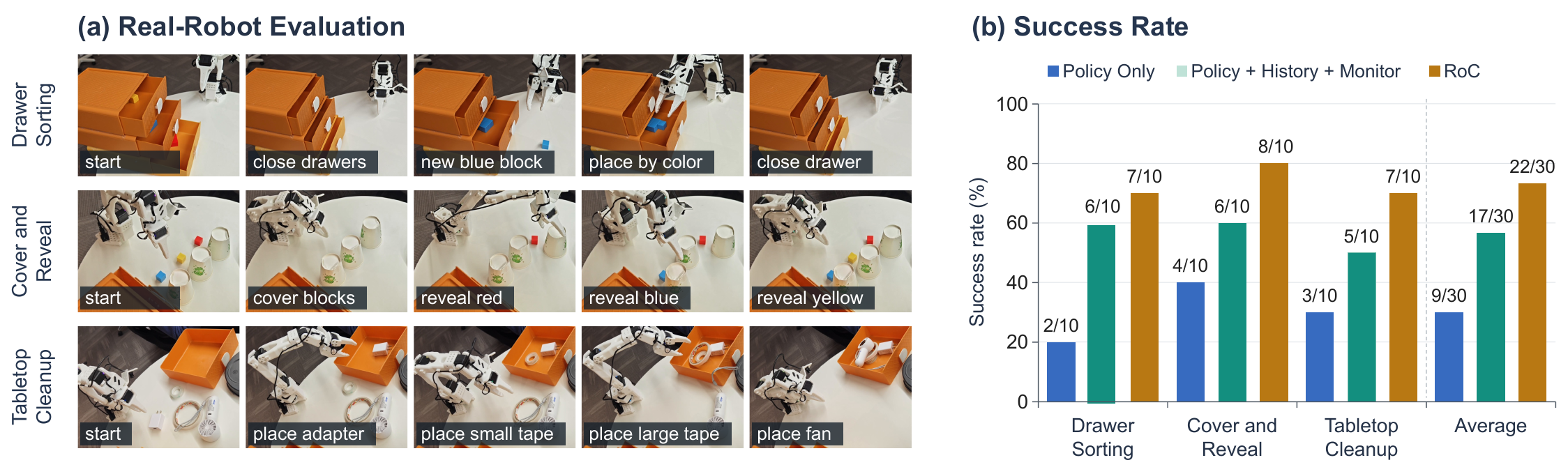}
      \vspace{-0.5cm}
  \caption{Real-robot evaluation on SO-ARM101. (a) Representative sequences for Drawer Sorting, Cover and Reveal, and Tabletop Cleanup. (b) Success over 10 trials per task.}
    \vspace{-0.4cm}
  \label{fig:real_robot}
\end{figure*}

\paragraph{Tasks.}
We evaluate three tabletop manipulation tasks on the real robot, illustrated in Figure~\ref{fig:real_robot}(a). In Drawer Sorting, the robot observes the initial color-to-drawer associations and then closes the drawers. It is then given a randomly selected yellow, blue, or red block, which it places in the drawer at the corresponding level. In Cover and Reveal, the robot memorizes the randomized initial locations of three colored blocks, covers them with cups from left to right, and then reveals the red, blue, and yellow blocks in that order. In Tabletop Cleanup, the robot places a white power adapter, a small transparent tape roll, a large white tape roll, and a white handheld fan into the orange box.

\paragraph{Results and Analysis.}
Figure~\ref{fig:real_robot}(b) reports success over 10 trials per task for three variants of RoC: Policy Only, Policy + History + Monitor, and RoC. RoC achieves success rates of 70\%, 80\%, and 70\% on Drawer Sorting, Cover and Reveal, and Tabletop Cleanup, respectively. Its overall success rate is 73.3\%, compared with 56.7\% for Policy + History + Monitor and 30.0\% for Policy Only. Adding history and monitoring improves task progression, but the resulting system still relies on the policy to recover from unsuccessful interactions. Agent intervention yields additional gains of 10, 20, and 20 percentage points across the three tasks, respectively. These results show how selective agent intervention complements task-state monitoring by enabling execution repair and plan revision before errors propagate to later stages. Further efficiency comparisons with agent-based methods are provided in the appendix.

\paragraph{Failure Analysis.}
Failures primarily involve difficult grasps that remain unsuccessful despite agent intervention. In these cases, the current reasoning agent fails to translate visual evidence into effective grasp corrections, highlighting agent capability as a remaining bottleneck. RoC decouples agent reasoning from task-state monitoring and policy execution, allowing more capable agents to be integrated without redesigning the control framework. Stronger visually grounded manipulation reasoning may therefore improve recovery within the same architecture.



\section{Conclusion}
\label{sec:conclusion}
In this work, we presented RoC, an efficient agentic control framework for long-horizon robot manipulation. RoC coordinates a low-level policy, an online task-state monitor, and a reasoning agent to enable selective agent intervention and coordinated execution repair and plan revision. By tracking subtask progress and verifying execution outcomes, RoC allows routine manipulation to remain with the policy while invoking higher-level reasoning when needed. Experiments on RMBench demonstrate improved task success over representative agent-based and policy-based methods, together with reduced token consumption relative to the agent-based baselines. Real-world experiments further illustrate the effectiveness of timely intervention in handling execution errors and incorrect subtask transitions. These findings highlight the importance of explicit task-state awareness and selective reasoning for efficient and reliable long-horizon robot manipulation.

\section*{LLM Usage Statement}
OpenAI Codex was used to assist with manuscript editing,
related-work research, and review of the controller code.
The authors reviewed the resulting suggestions.
No experimental measurements were generated using these tools.

{\small
\bibliographystyle{ieeenat_fullname}
\bibliography{references}
}
\end{document}